\documentclass[11pt,a4paper]{article}
\usepackage[hyperref]{ranlp2025}
\usepackage{times}
\usepackage{latexsym}

\usepackage{microtype}

\usepackage{graphicx}
\usepackage{amsmath}
\usepackage{booktabs}
\usepackage{subcaption}  
\usepackage[T1]{fontenc} 

\aclfinalcopy 

\title{Multilingual != Multicultural: Evaluating Gaps Between Multilingual Capabilities and Cultural Alignment in LLMs}

\author{Jonathan Rystrøm\\
  Oxford Internet Institute \\
  University of Oxford, UK \\\And
  Hannah Rose Kirk \\
  Oxford Internet Institute \\
  University of Oxford, UK \\
  Correspondence: \texttt{jonathan.rystrom@oii.ox.ac.uk}\\\And
  Scott A. Hale \\
  Oxford Internet Institute\\
  University of Oxford, UK \\}

\date{}

\begin{document}
\maketitle
\begin{abstract}
Large Language Models (LLMs) are becoming increasingly capable across global languages. However, the ability to communicate across languages does not necessarily translate to appropriate cultural representations. A key concern is US-centric bias, where LLMs reflect US rather than local cultural values. We propose a novel methodology that compares LLM-generated response distributions against population-level opinion data from the World Value Survey across four languages (Danish, Dutch, English, and Portuguese). Using a rigorous linear mixed-effects regression framework, we compare three families of models: Google's Gemma models (2B--27B parameters), AI2's OLMo models (7B-32B parameters), and successive iterations of OpenAI's turbo-series. Across the families of models, we find no consistent relationships between language capabilities and cultural alignment. While the Gemma models have a positive correlation between language capability and cultural alignment across all languages, the OpenAI and OLMo models are inconsistent. Our results demonstrate that achieving meaningful cultural alignment requires dedicated effort beyond improving general language capabilities.
\end{abstract}



\section{Introduction}
\begin{figure}
    \centering
    \includegraphics[width=0.9\linewidth]{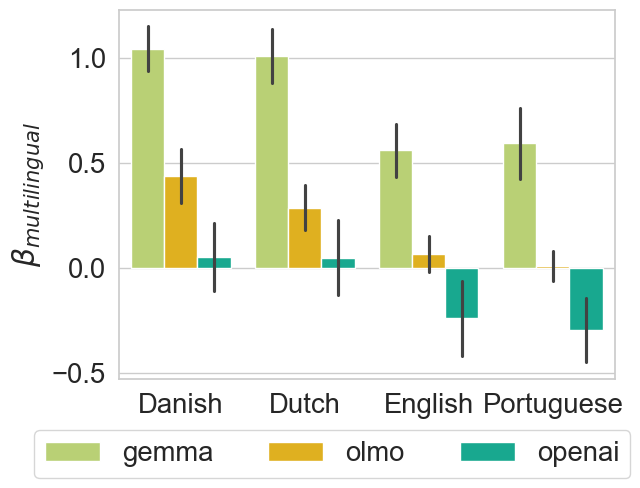}
    \caption{The relationship between multilingual capability and cultural alignment is inconsistent across LLM families, as shown by coefficients from our linear mixed-effects model ($\beta_{multilingual}=\beta_{flm}$; Eq.~\ref{eq:rq1-eq}; §\ref{sec:methods-rq1}). OpenAI and OLMo models show negative or insignificant relationships outside of Danish and Dutch, while Gemma models show positive relationships throughout ($p<.05$).}
    \label{fig:rq1-eq-main}
\end{figure}
Spearheaded by accessible chat interfaces to powerful models like ChatGPT \citep{openaiChatGPTOptimizingLanguage2022}, LLMs are reaching hundreds of millions of users \citep{milmoChatGPTReaches1002023}. These models are deployed across diverse contexts: from tutoring mathematics \citep{khanHarnessingGPT4That2023} to building software applications \citep{pengImpactAIDeveloper2023} to assisting in legal cases \citep{tanChatgptArtificialLawyer2023}. While most LLMs demonstrate multilingual abilities \citep{ustunAyaModelInstruction2024}, the ability to communicate across languages does not necessarily translate into appropriate cultural representations. Disentangling language capabilities and cultural alignment is crucial for understanding how LLMs should be examined and audited \citep{mokanderAuditingLargeLanguage2024} and for ensuring these technologies work for diverse people \citep{dignazioDataFeminism2023,weidingerTaxonomyRisksPosed2022}.

Given the Silicon Valley origins of many frontier AI labs and the prevalence of American English training data, we might expect LLMs to exhibit US-centric cultural biases despite their multilingual capabilities. These companies comprise a narrow slice of human experience, limiting the voices that contribute to critical design decisions in LLMs \citep{dignazioDataFeminism2023}. They typically train LLMs on massive amounts of predominantly English text and employ American crowd workers to rate and evaluate the LLMs' responses \citep{johnsonGhostMachineHas2022,kirkPresentBetterFuture2023}. Far too often, the benefits and harms of data technologies are unequally distributed, reinforcing biases and harming already minoritized groups \citep{birhaneAlgorithmicColonizationAfrica2020,milanBigDataSouths2019,khandelwalIndianBhEDDatasetMeasuring2024}. Understanding how LLMs represent different cultures is thus paramount to establishing risks of representational harm \citep{rauhCharacteristicsHarmfulText2022} and ensuring the technology's utility is shared across diverse communities.

Increasing diversity and cross-cultural understanding is stymied by unchecked assumptions in both alignment techniques and evaluation methodologies. First, there is an assumption that bigger and more capable LLMs trained on more data will be inherently easier to align \citep{zhouLIMALessMore2023, kunduSpecificGeneralPrinciples2023}, but this sidesteps the thorny question of pluralistic variation and cultural representations \citep{kirkPRISMAlignmentDataset2024}. Thus, it is unclear whether improvements in architecture \citep{fedusReviewSparseExpert2022} and post-training methods \citep{kirkPresentBetterFuture2023,rafailovDirectPreferenceOptimization2023a} translate into improvements in cultural alignment.

Although studies like the World Values Survey (WVS) have documented how values vary across cultures \citep{evs/wvsJointEVSWVS2022}, it remains unclear whether more capable LLMs---through scaling or improved training---better align with these cultural differences \citep{baiTrainingHelpfulHarmless2022,kirkPresentBetterFuture2023}. While the WVS has been used in prior research on values in LLMs, these studies have focused predominantly on individual models' performance within an English-language context. \citep{caoAssessingCrossculturalAlignment2023,aroraProbingPretrainedLanguage2023,alkhamissiInvestigatingCulturalAlignment2024}. This paper addresses this gap by developing a methodology for assessing how well families of LLMs represent different cultural contexts across multiple languages. We compare two distinct paths to model improvement: systematic scaling of instruction-tuned models and commercial product development comprising scaling and innovation in post-training to accommodate pressures from capabilities, cost, and preferences \citep{openaiGPT4oSystemCard2024}.

Given these considerations, we investigate the following research questions:

\begin{description}
   \item[RQ1] \textbf{Multilingual Cultural Alignment:} Does improved multilingual capability increase LLM alignment with population-specific value distributions?
   \item[RQ2] \textbf{US-centric Bias:} When using different languages, do LLMs align more with US values or with values from the countries where these languages are native?
\end{description}

We operationalise \textit{multilingual capability} as an LLM's performance on a range of multilingual benchmarks across languages \citep[see, e.g.,][]{nielsenScandEvalBenchmarkScandinavian2023a}. We describe the specific benchmarks and performances in the \href{https://github.com/jhrystrom/multicultural-alignment/blob/main/supplementary/multilingual_capability_table.md}{supplementary materials}.

This work makes several key contributions. First, we introduce a novel distribution-based methodology for probing cultural alignment across languages, moving beyond direct survey approaches to better capture latent cultural values \citep{sorensenPositionRoadmapPluralistic2024}. Second, we provide the first systematic comparison of how improvements in scale and post-training affect cultural alignment and US-centric bias across English, Danish, Dutch, and Portuguese through a series of robust statistical models. Third, we release a dataset of model-generated responses across multiple languages and cultural contexts as well as our code, enabling future research into cultural alignment and bias.\footnote{See \href{https://github.com/jhrystrom/multicultural-alignment}{github.com/jhrystrom/multicultural-alignment} for code, data, and supplementary materials.} Together, these contributions advance our understanding of how LLM development choices influence cultural representation while providing tools for ongoing investigation of these critical issues.

\section{Measuring Cultural Alignment} \label{sec:measure-cultural}
\begin{figure}[ht]
    \centering
    \includegraphics[width=\linewidth]{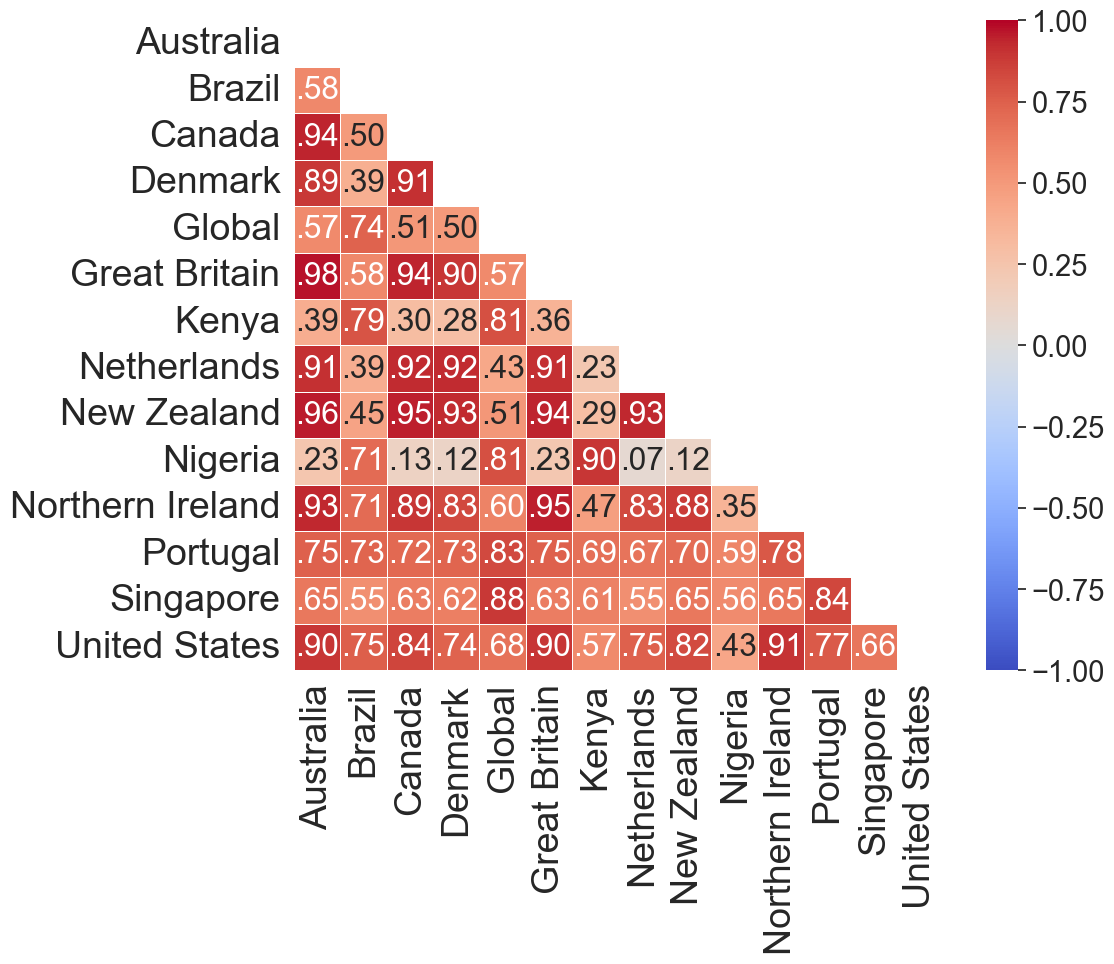}
    \caption{Pearson correlations in value polarity scores across studied countries from the World Values Survey. Value polarity scores are the fraction of the population in favour of a given topic. All correlations are positive, with most being between 0.7--0.95.}
    \label{fig:wvs-correlations}
\end{figure}

This section defines `cultural alignment' and how to measure it in LLMs. We conceptualise cultural alignment as reproducing distributions of values in a particular population. Then we show how to a) get a ground-truth distribution of values using the World Values Survey (§\ref{sec:wvs}) and b) elicit value distributions from LLMs (§\ref{sec:llm-responses}).

\paragraph{Cultural alignment as value reproduction:}
Within a culture there will be a variety of stances to any particular topic. However, the \textit{distribution} of stances will be characteristic among cultures. For instance, while around 8\% of Danes are opposed to abortion, it is a much less contentious topic than in the US, where it's close to 40\% \cite{evs/wvsJointEVSWVS2022}. 

We posit that cultural alignment for a specific group of people can be operationalised as how well an LLM reproduces the distribution of values over a wide range of topics \citep{sorensenPositionRoadmapPluralistic2024}. Investigating \textit{distributions} of responses differs from previous work that directly surveys the LLMs as regular participants \cite[e.g.,][]{caoAssessingCrossculturalAlignment2023}. This approach also addresses concerns raised by \citet{khanRandomnessNotRepresentation2025} about the instability of survey-based evaluations by focusing on aggregate distributions rather than individual responses and incorporating explicit controls for response consistency. Our goal is to get more naturalistic elicitations of the underlying values whilst avoiding sycophancy and response bias \cite{sharmaUnderstandingSycophancyLanguage2023}. 

We operationalise reproduction as high correlations between \textit{value polarity scores}: the fraction of people (or LLM responses) in favour of a topic in the population. Note, that we binarise issues to allow for simpler operationalisation. Below, we describe how we empirically estimate the value polarity score for the ground truth (§\ref{sec:wvs}) and LLMs (§\ref{sec:llm-responses}). 

\subsection{Ground Truth: World Values Survey} \label{sec:wvs}
To get a `ground truth' distribution of cultural values, we use the joint World Values Survey and European Values Survey \citep[EVS;][]{evs/wvsJointEVSWVS2022}. These surveys cover adults across 92 countries with samples that are nationally representative for gender, age, education, and religion. The surveys' broad coverage enables cross-cultural comparability for the many countries covered by the surveys, though some scholars note challenges in ensuring response comparability across countries \citep{alemanValueOrientationsWorld2016}. The WVS provides both country and language identifiers for each respondent, allowing us to define populations either as citizens of a country or speakers of a language using the same underlying respondent-level data.

We select questions with binary agree/disagree or rating scale formats that allow clear classification of positive vs. negative stances, excluding questions with multiple categorical response options (see the \href{https://github.com/jhrystrom/multicultural-alignment/blob/main/supplementary/wvs_all_topics.md}{supplementary materials} for the full list of questions). These questions span environment, work, family, politics, religion, and security. We convert responses to binary indicators by determining whether each response indicates support for the measured construct, with custom coding to handle the various question formats and reverse-scored items. Finally, we calculate the value polarity score as the demographically weighted proportion of respondents with affirmative stances. Formally, we can define the value polarity score for a given population, $\mathcal{P}$ (e.g., citizens in a country or speakers of a language) and topic, $q$, (i.e., question within the EVS/WVS) as shown in Eq. \ref{eq:vps-human}: 

\begin{equation} \label{eq:vps-human}
\mathrm{VPS}_{\mathcal{P},q} 
= \sum_{i \in \mathcal{P}} 
\frac{w_i}{\sum_{j \in \mathcal{P}_q} w_j} \, A_{i,q}
\end{equation}

Here, $A_{i,q}$ is a binary indicator of whether participant $i$ has a positive stance on topic $q$, $w_i$ represents the survey-provided demographic weights, and $\mathcal{P}_q$ denotes respondents in population $\mathcal{P}$ who answered question $q$. The first term normalises the weights to account for missing responses and enables aggregation across any definition of a population (e.g., residents in a country, speakers of a language, etc.).

For example, if 80\% of Danish respondents who answered the same-sex marriage question expressed support (after demographic reweighting), Denmark's value polarity score for this topic would be 0.8. Thus, a culture's values can be represented as a vector, where each element corresponds to a value polarity score for a specific topic.

\subsection{Ecologically valid LLM responses} \label{sec:llm-responses}
Testing cultural alignment effectively requires embedding contextual and cultural elements in ways that maintain ecological validity. At a high level, eliciting values from an LLM consist of two steps: 1) Iteratively prompting the model with the selected topics and 2) extracting the stances from each model response.

\paragraph{Setting prompt context:}
Developing ecologically valid prompts requires careful consideration. When evaluating LLM responses to value-laden topics, simply asking questions like ``What proportion of people support {topic X}?'' or ``Do you support {topic X}?'' proves inadequate \citep[e.g.,][]{rozadoPoliticalPreferencesLLMs2024}. Such direct approaches suffer from three key limitations: they generate false positives through excessive agreement, fail to reflect realistic usage patterns, and provide insufficient variation to assess cultural alignment \citep{rottgerPoliticalCompassSpinning2024}. They also struggle to capture instance-specific harms that emerge when systems misalign with users' cultural contexts \citep{rauhCharacteristicsHarmfulText2022}.

Instead, we adopt an implicit approach by asking the model to generate responses from hypothetical respondents. For example, prompting ``imagine surveying 10 random people on {topic X}. What are their responses?'' This method reveals the model's latent opinion distribution while avoiding the limitations of direct questioning. Details for prompt construction are provided in the \href{https://github.com/jhrystrom/multicultural-alignment/blob/main/supplementary/prompt\_construction.md}{supplementary materials}.

\paragraph{Seeding cultural responses:}
Having a method for eliciting distributions of values, the next step is to seed culture. One typical way of seeding a specific culture is to explicitly instruct the LLM either by mentioning a specific country (`imagine surveying 10 random Americans') or through describing specific personas \citep[`Imagine surveying a 85-year-old Danish woman...';][]{alkhamissiInvestigatingCulturalAlignment2024}. The problem with these demographic prompting approaches is that they stray from actual uses of LLMs. Users are unlikely to explicitly mention their demographic information or nationality \citep{zhengLMSYSchat1MLargescaleRealworld2023}. 

Instead, we use language as a proxy for cultural origin. For instance, a prompt in Danish is assumed to come from a Dane. This approach creates an intentional distinction in our analysis: we can compare `language-level' alignment (all speakers of a language globally) with `country-level' alignment (all people from specific nations where that language is native). As argued by \citet{havaldarMultilingualLanguageModels2023}, users speaking a particular language would expect culturally appropriate responses in that language. For languages spoken in multiple countries, this approach is intentionally ambiguous. The ambiguity allows us to elicit the underlying `default' alignment rather than the general ability to emulate cultures \citep{taoCulturalBiasCultural2024}. We validate this approach by showing that LLM responses exhibit significantly lower self-consistency between languages compared to within languages, demonstrating that language impacts output (see the \href{https://github.com/jhrystrom/multicultural-alignment/blob/main/supplementary/language_conditioning.md}{supplementary materials}). To create prompts across languages, we use \texttt{gpt-3.5-turbo} to translate our original English prompts. Although previous literature has shown strong translation capabilities in LLMs \citep{yanGPT4VsHuman2024}, we nonetheless manually verify the translations.


\paragraph{Annotating and aggregating responses:}
Finally, to transform the LLMs' hypothetical survey responses into vectors of stances, we use an LLM-as-a-judge approach \cite{zhengJudgingLLMasajudgeMTbench2023,guerdanValidatingLLMasajudgeSystems2025}. Specifically, we use \texttt{gpt-4.1-mini} \citep{openaiIntroducingGPT412025} to label each substatement as either `pro', `con', or `null' given the context of the topic and a representative pro and con statement (generated with an LLM and validated by the authors). We then calculate the proportion of `pro' versus `con' responses as the LLM's value polarity score for the given statement. For instance, a response with seven `pro', one `con', and two `null' statement would yield a value polarity score of 0.875 ($\frac{7}{8}$). A complete, unabridged example can be found in the \href{https://github.com/jhrystrom/multicultural-alignment/blob/main/supplementary/response_example.md}{supplementary materials}. Formally, we label each substatement from the full set of hypothetical statements, $G_{q,g}$, for topic $q$ and generation $g$ as $r$. Furthermore, we label the classifier as $\ell(r)$. We then formalise the value polarity score for a given instance of a generation for a topic ($\mathrm{VPS}^{\mathrm{LLM}}_{q,g}$) as shown in Eq. \ref{eq:vps-llm}: 

\begin{equation} \label{eq:vps-llm}
\mathrm{VPS}^{\mathrm{LLM}}_{q,g}
= \frac{\sum_{r \in G_{q,g}} [\,\ell(r) = \mathrm{pro}\,]}
       {\sum_{r \in G_{q,g}} [\,\ell(r) \in \{\mathrm{pro},\mathrm{con}\}\,]},
\end{equation}

These scores are then compared against the value polarity scores from the WVS. Specifically, we calculate the Spearman rank correlation to obtain a measure of similarity between the LLMs' responses and the value distributions of a given population. 

To validate the LLM-as-judge, we manually annotate 200 statements. We iteratively refine the prompts and the LLM used until we reach satisfactory performance. We find a 91\% agreement and a mean absolute error for value polarity of 4.5\% over the dataset, ensuring consistent statistics between LLM and human annotation \citep{guerdanValidatingLLMasajudgeSystems2025}.

\section{Experimental Setup}
\begin{figure}
    \centering
    \includegraphics[width=\linewidth]{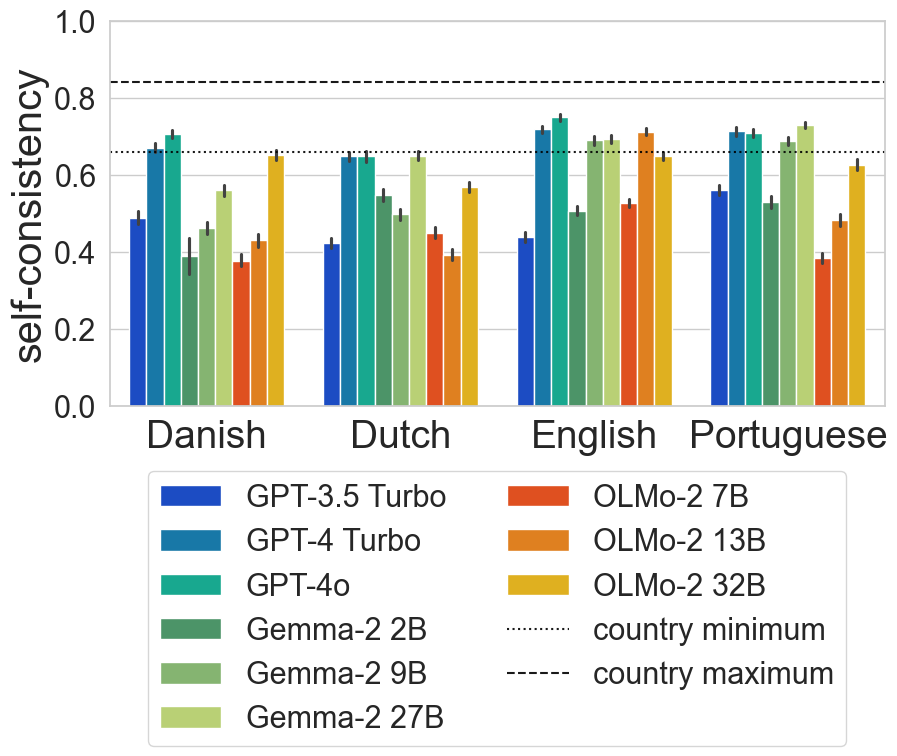}
    \caption{Self-consistency in responses for LLMs and WVS countries. LLMs have lower self-consistency than resampled WVS responses---shown by the dashed lines---particularly in non-English languages.}
    \label{fig:self-consistency}
\end{figure}
To investigate whether improving the multilingual capabilities of LLMs improves cultural alignment, we set up an experiment using a carefully chosen set of models and languages. We examine two different kinds of model improvements: scaling and commercial product development. These cases provide complementary perspectives on the effects of multilingual capabilities on cultural alignment. Scaling is the most well-studied path to improving LLMs \citep{kaplanScalingLawsNeural2020,ganguliPredictabilitySurpriseLarge2022}. Commercial product development, on the other hand, comprises both scale and innovation in post-training to accommodate different pressures from capabilities, cost, and preferences \citep{kirkBenefitsRisksBounds2024}. For scaling, we use the instruction-tuned Gemma models \cite{gemmaGemma2Improving2024} and OLMo-2 models \cite{olmo2OLMo22025}, while for product development, we use OpenAI's turbo-series models \cite{openaiChatGPTOptimizingLanguage2022,openaiGPT4TechnicalReport2024,openaiGPT4oSystemCard2024}. We provide details of these model families in §\ref{models}. A breakdown of the computational cost is in the \href{https://github.com/jhrystrom/multicultural-alignment/blob/main/supplementary/experiment_cost.md}{supplementary materials}.

\paragraph{Languages:}

For the languages, we compare English with Danish, Dutch, and Portuguese. This set allows us to test multiple assumptions about cultural alignment. English represents a widely used case: it is a global language with speakers across many countries represented in the WVS (see Fig. \ref{fig:wvs-correlations}). This diversity allows us to assess whether LLMs align more strongly with US values or those of other English-speaking nations.

Danish and Dutch serve as controlled test cases since they are primarily used in a single country. If cultural alignment stems from pre-training data, models should show strong Danish/Dutch cultural alignment when using these languages, despite their small share of training data \citep{kreutzerQualityGlanceAudit2022}. Alternatively, if alignment emerges from post-training processes---which are predominantly English-based \citep{blevinsLanguageContaminationHelps2022}--responses in these languages should align more with US values.

Portuguese presents an interesting case since it is an official language in several countries. We investigate whether the LLM responses are more aligned to Portugal or Brazil---two countries that show distinct value patterns in relation to each other and the US (see Fig. \ref{fig:wvs-correlations}). This allows us to test whether an LLM aligns more strongly with one country's values, the aggregate values of all language users, or US values.

For each language-model pair, we collect 300 prompt-response pairs to power our statistical analysis sufficiently (see §\ref{sec:methods-rq1}). After filtering out responses that either lacked the required hypothetical survey format or were in a language other than the prompt, we obtained between 111--299 valid responses per combination. We calculate the correlation in value polarity scores at three levels: country (e.g., US or Denmark), language (pooling all speakers of a given language), and global (weighted values from all WVS/EVS participants). 

\subsection{Models} \label{models}
We examine three model families representing different development approaches: Gemma \cite{gemmaGemma2Improving2024} and OLMo \citep{olmo2OLMo22025} for improvements through scaling and OpenAI's turbo series for commercial product development, combining scaling with post-training improvements \citep{openaiChatGPTOptimizingLanguage2022,openaiGPT4TechnicalReport2024,openaiGPT4oSystemCard2024}. Other preliminary experiments included different versions of LLaMA models \citep{touvronLLaMAOpenEfficient2023} and Mistral models \citep{jiangMistral7B2023}. However, these models either failed to consistently follow instructions or always answered in English regardless of the prompt language. See the \href{https://github.com/jhrystrom/multicultural-alignment/blob/main/supplementary/llm_descriptions.md}{supplementary materials} for a more thorough description of the LLMs.

\subsection{RQ1: Multilingual Cultural Alignment} \label{sec:methods-rq1}
\begin{figure}[htbp]
    \centering
    \includegraphics[width=0.9\linewidth]{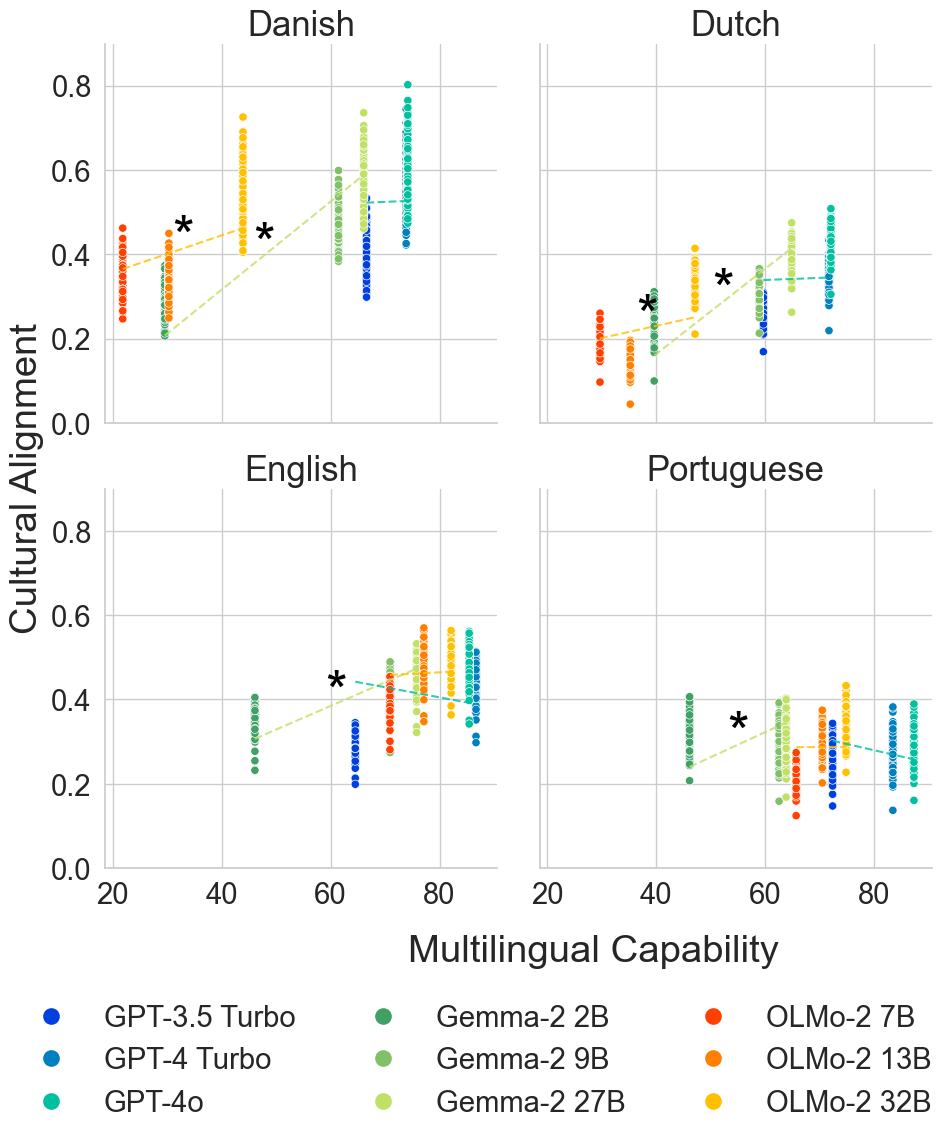}
    \caption{Language capability (x-axis) vs cultural alignment scores (y-axis) across languages. Stars indicate significance ($p<.05$) in our linear mixed-effects regression of multiple runs (See §\ref{sec:methods-rq1}). OpenAI models (blue) and OLMo models (red) show negative/insignificant relationships outside of English, while the Gemma models (green) show positive relationships throughout ($p<.05$).}
    \label{fig:multilingual-alignment-scatter}
\end{figure}

To statistically assess whether improving the multilingual capabilities of LLMs improves cultural alignment, we construct a linear mixed-effects regression \citep[LMER;][]{lukeEvaluatingSignificanceLinear2017} based on the experimental setup described above. Our LMER follows standard practices and has three core components: 

\begin{itemize}
    \item \textbf{Core coefficient:} The coefficient of interest is the three-way interaction between model family, language, and multilingual capability. This tests whether the multilingual capability–alignment relationship differs by model family and response language, directly addressing \textbf{RQ1}.
    \item \textbf{Random effects:} We include a model-specific random intercept $\alpha_j$ to account for repeated measures of cultural alignment for the same LLM. This models variation between LLMs and can improve efficiency over standard linear regressions \citep{lukeEvaluatingSignificanceLinear2017}.
    \item \textbf{Control for self-consistency:} We include a consistency-by-language term to help ensure that higher alignment scores reflect genuine cultural adaptation rather than reduced response noise, which can inflate scores \citep{kahnemanNoiseFlawHuman2021}.
\end{itemize}

We calculate self-consistency as the Spearman correlation between value polarity scores (defined in §\ref{sec:measure-cultural}) of repeated responses to identical topics, adjusted by the reliability of the LLM annotation \citep[see §\ref{sec:llm-responses};][]{charlesCorrectionAttenuationDue2005}. A score of 1.0 indicates perfect consistency; 0.0 indicates random responses. Population-level resampling of the human WVS responses yields values between 0.66 and 0.84 (see Fig.\ref{fig:self-consistency} and the \href{https://github.com/jhrystrom/multicultural-alignment/blob/main/supplementary/wvs_all_topics.md}{supplementary materials}).

Formally, the model is specified in Eq. \ref{eq:rq1-eq}:

\begin{equation} \label{eq:rq1-eq}
\begin{split}
\text{CA}_i &\sim \mathcal{N}(\mu_i, \sigma^2), \\
\mu_i &= \alpha_{j[i]} + \beta_{1l} \, X_{\text{cons},i} X_{l,i} \\
    &\quad + \beta_{flm} \, X_{m,i} X_{f,i} X_{l,i}, \\
\alpha_j &\sim \mathcal{N}(\mu_\alpha, \sigma_\alpha^2), \quad j = 1, \ldots, J.
\end{split}
\end{equation}

where $i$ indexes responses and $j[i]$ denotes the LLM producing response $i$. Here $X_{\text{cons},i}$ is the self-consistency score for response $i$, $X_{l,i}$ is the set of language indicators, $X_{f,i}$ is the set of model-family indicators, and $X_{m,i}$ is the multilingual capability score. The residual variance $\sigma^2$ represents within-LLM variation in alignment scores not explained by the fixed effects or model-specific intercept, while $\sigma_\alpha^2$ represents between-LLM variation in average alignment.

The above statistical model allows us to analyse the relationship between multilingual capabilities and cultural alignment in model families at the level of individual languages. For example, we might find that multilingual capabilities improve cultural alignment for Gemma models for Danish but not for Dutch or vice versa. 

\subsection{RQ2: US-Centric Bias} \label{methods-rq2}
We analyse model bias by comparing cultural alignment between US and local values, where ``local'' refers to values in the country or countries where a given language is natively spoken. We define US-centric bias as an LLM showing higher cultural alignment with US value distributions compared to local ones. To quantify this bias, we use a linear regression model that measures the differential effect of US versus local value alignment:

\begin{equation} \label{eq:rq2-eq}
\begin{aligned}
\operatorname{CA} &= \beta_{0} + \beta_1(\operatorname{US}) \\
&+ \sum_{m \in \mathcal{M}} \sum_{l \in \mathcal{L}} \beta_{ml}(m \times l) \\
&+ \sum_{m \in \mathcal{M}} \sum_{l \in \mathcal{L}} \beta_{ml}^{\operatorname{US}}(\operatorname{US} \times m \times l) + \epsilon
\end{aligned}
\end{equation}

The regression's intercept ($\beta_0$,i.e., the base case) is a baseline that produces uniformly random value polarity scores. $\mathcal{M}$ is the set of models and $\mathcal{L}$ is the set of languages. \texttt{US} is a boolean feature denoting whether the cultural alignment is to the US (if $1$) or the local values (if $0$). We primarily analyse the coefficients with US ($\beta_{ml}^{US}$) since these provide the \textit{partial} effect of US-centric bias, i.e., how much more/less a given LLM is aligned to US rather than local values. Assumption checks for the regression can be seen in the \href{https://github.com/jhrystrom/multicultural-alignment/blob/main/supplementary/assumption_check.md}{supplementary materials}.

\section{Results}

\subsection{Multilingual Cultural Alignment (RQ1)} \label{sec:results-improvements-rq1}
We first examine the stability of LLMs' cultural values. For LLMs lacking stable internal values, apparent improvements in cultural alignment may reflect reduced response variance rather than genuine advances \citep{rottgerPoliticalCompassSpinning2024,kahnemanNoiseFlawHuman2021}. We therefore analyse both the self-consistency of LLM responses and how alignment changes with model improvements.

\paragraph{LLMs have low self-consistency:}
We find low self-consistency scores across all models and languages compared to human responses in the WVS data (Fig. \ref{fig:self-consistency}). In contrast, LLMs show generally lower self-consistency compared to the human responses, even in English, where instruction-following capabilities are strongest due to English-dominated training data. \citep{openaiGPT4TechnicalReport2024,gemmaGemma2Improving2024,olmo2OLMo22025}.

This lower self-consistency complicates our cultural alignment analysis \cite{wrightLLMTropesRevealing2024}. Drawing on \citet{kahnemanNoiseFlawHuman2021}'s noise framework, we recognise that inconsistent responses can be as detrimental as bias with respect to the accuracy of the analysis. To address the noise, we employ larger sample sizes and incorporate consistency controls in our regression analyses.

\paragraph{Multilinguality does not imply cultural alignment:}
The relationship between model improvements and cultural alignment varies substantially across languages and model families (Fig. \ref{fig:rq1-eq-main}). For Gemma, there is a strong and significant positive relationship between multilingual capabilities and cultural alignment for all languages. In contrast, the relationships for the GPT-Turbo models are either insignificant or negative. For Dutch and Danish the relationships are insignificant ($\beta_{\text{gpt},\text{nl}}=0.049, p=0.589$,$\beta_{\text{gpt},\text{da}}=0.053, p=0.522$), and for Portuguese and English the effect is significant and negative ($\beta_{\text{gpt},\text{en}}=-0.24,p=0.009$, $\beta_{\text{gpt},\text{pt}}=-0.30,p<0.001$). Similarly for OLMo, the relationship is positive for Danish and Dutch ($\beta_{\text{OLMo},\text{da}}=0.44,p<0.001$, $\beta_{\text{OLMo},\text{nl}}=0.29,p<0.001$) and insignificant for English and Portuguese ($\beta_{\text{OLMo},\text{en}}=0.068,p=0.115$, $\beta_{\text{OLMo},\text{pt}}=0.008,p=0.825$).

The mismatch between multilingual performance and cultural alignment could suggest a capability threshold: multilingual improvements might provide rudimentary instruction following skills \cite{nieMultilingualLargeLanguage2024}, but beyond a point, other factors---such as the preferences of developers and annotators---dominate \citep{kirkPRISMAlignmentDataset2024}. This could explain the smaller open weights models' higher coefficients than the gpt-turbo models (see Fig. \ref{fig:multilingual-alignment-scatter} or Fig. \ref{fig:rq1-eq-main}). Further work is needed to understand alignment at the sub-national level.

Furthermore, the strong effect of self-consistency ($0.405<\beta_{\text{consistency}}<0.723, p \ll 0.001$) compared to multilingual capability suggests that noise remains a major limiting factor in analysing cultural alignment. This aligns with broader findings about the instability of LLM value elicitation \citep{rottgerPoliticalCompassSpinning2024,khanRandomnessNotRepresentation2025}. Moreover, even the highest observed alignment scores (around 0.7; see Fig \ref{fig:multilingual-alignment-scatter}) indicate substantial room for improvement in how well LLMs match human cultural values and behaviours.

In conclusion, our analysis reveals a complex relationship between model improvements and cultural alignment. Although some languages show progressive improvements in cultural alignment from model scaling or iterative commercial development, others show minimal or inconsistent improvements. These findings, combined with the relatively low self-consistency of LLM responses, demonstrate that improved multilingual capability does not guarantee better cultural alignment.

\subsection{US-centric Bias (RQ2)}
\begin{figure}
    \centering
    \includegraphics[width=\linewidth]{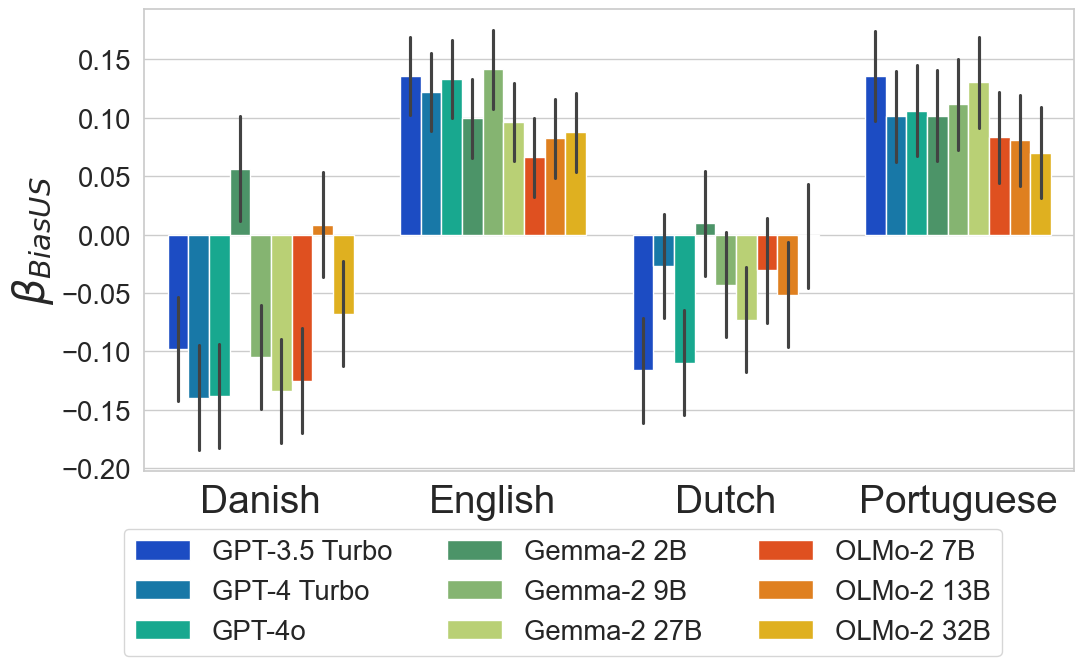}
    \caption{US-centric bias coefficients across LLMs and languages ($\beta_{BiasUS}$); see Eq. \ref{eq:rq2-eq}). Error bars are standard errors from the regression. Positive values indicate the presence of US-centric bias.}
    \label{fig:rq2-us-bias}
\end{figure}

Here, we answer RQ2 by examining US bias across languages. Specifically, we investigate relative alignment between local and US values (Fig. \ref{fig:rq2-us-bias}).

Our analysis reveals distinct patterns of US-centric bias across both languages and model families (Fig. \ref{fig:rq2-us-bias}). Languages show different susceptibilities to US bias: only one of nine LLMs exhibits US-centric bias in Danish, all in English, all in Portuguese, and none in Dutch. Note that for English, these results mean that the LLM, on average, is relatively more aligned to US values compared to other English-speaking countries like Kenya or the United Kingdom. See the \href{https://github.com/jhrystrom/multicultural-alignment/blob/main/supplementary/language_breakdown.md}{supplementary materials} for detailed results.

The overarching pattern is that languages spoken \textit{across} countries (English and Portuguese) show US-centric bias, whereas languages spoken in only \textit{one} country (Danish and Dutch) show less US-centric bias. This supports the hypothesis that homogeneity in the training data can counteract US-centric bias---at least for medium-resourced, Western-European languages. 

For LLMs, some specific LLMs seem more prone to bias across languages. Specifically, the small \texttt{gemma-2-2b-it} exhibits higher US-centric bias across every language except Dutch. Beyond that, we see no clear progressions in US-centric bias within any family.

In conclusion, language seems a stronger indicator of US-centric bias in LLMs compared to LLM development. Monocultural languages show insignificant to negative bias, while English and Portuguese show significant US-centric bias. Within each LLM family, we find no consistent nor significant change in US-centric bias across LLM versions. These findings underscore the complex relationship between multilingual capability and alignment.

\section{Related Work}
Recent work emphasizes the need for systematic auditing of LLMs' cultural alignment, particularly as these models are deployed globally \citep{kirkBenefitsRisksBounds2024, mokanderAuditingLargeLanguage2024,kirkPRISMAlignmentDataset2024}. Prior empirical approaches have primarily taken two paths: using transformations based on Hofstede's cultural dimensions framework or directly comparing against survey responses. Studies using Hofstede's dimensions \citep{masoudCulturalAlignmentLarge2025, caoAssessingCrossculturalAlignment2023} provide structured cross-cultural comparisons through latent variable analysis. However, these studies assume that LLMs' latent dimensions map directly onto human dimensions, since they use formulas calibrated for humans---an assumption that warrants scrutiny \citep{shanahanTalkingLargeLanguage2024,schroderLargeLanguageModels2025}.

Recent work has explored using LLMs to simulate responses for assessing cultural alignment \citep{taoCulturalBiasCultural2024,alkhamissiInvestigatingCulturalAlignment2024, havaldarMultilingualLanguageModels2023}. Similarly to our work, these works show that LLMs struggle to represent underrepresented personas \citep{alkhamissiInvestigatingCulturalAlignment2024} and emotions \citep{havaldarMultilingualLanguageModels2023} for non-English languages. Prior approaches focused on individual-level responses. In contrast, our method generates distributions of opinions across hypothetical survey participants, enabling direct comparison with population-level statistics. This distribution-based approach offers three key advantages. First, it better captures the inherent variation in cultural values within populations, paving the way for investigating distributional alignment \citep{sorensenPositionRoadmapPluralistic2024}. Second, it enables principled statistical comparison against large-scale survey data like the World Values Survey \citep{evs/wvsJointEVSWVS2022}. Finally, the framework is easy to extend to new languages by automatically translating the prompts. We detail our quantitative framework for measuring alignment with observed population distributions in §\ref{sec:measure-cultural}.

There is also an increasing body of work investigating political biases in LLMs \cite{rottgerPoliticalCompassSpinning2024,rottgerIssueBenchMillionsRealistic2025,rozadoPoliticalPreferencesLLMs2024}. Much of this work also relies on human political surveys like the Political Compass Test. However, recent work has called for increased attention to how the randomness inherent in LLM decoding at non-zero temperatures can create instability in attributes \cite{rottgerPoliticalCompassSpinning2024,wrightLLMTropesRevealing2024,khanRandomnessNotRepresentation2025}. We expand on this work by including multilingual perspectives and constructing prompts with a wide range of variations (see §\ref{sec:measure-cultural}). These prompt variations, combined with statistically accounting for self-consistency in our statistical analysis (see §\ref{sec:methods-rq1}), allow us to get a more robust measure of cultural alignment.

The relationship between model capabilities and cultural alignment remains understudied. Unlike general performance metrics that follow predictable scaling laws \citep{kaplanScalingLawsNeural2020}, cultural alignment may not improve systematically with model capabilities. This aligns with research showing micro-level capabilities can be discontinuous with scale \citep{ganguliPredictabilitySurpriseLarge2022}. The challenge is compounded in multilingual settings \citep{hoffmannEmpiricalAnalysisComputeoptimal2022}, where static benchmarks with single correct answers fail to capture how cultural values are distributed across different topics and contexts.

Previous work has focused primarily on English-language performance \citep{taoCulturalBiasCultural2024} or individual LLMs \citep{aroraProbingPretrainedLanguage2023,caoAssessingCrossculturalAlignment2023}. Our work extends this by examining how cultural alignment systematically varies within model families and across languages, providing insight into how different development approaches---scaling and commercial product development---influence cultural representation capabilities.

There is already progress on improving the cross-cultural participation in alignment data. Two notable projects are PRISM and AYA \citep{kirkPRISMAlignmentDataset2024,ustunAyaModelInstruction2024}. PRISM is a large dataset of conversational preferences from a diverse participant pool. While the data is predominantly in English, it could be an important resource for better understanding and modelling diverse cultural preferences. The AYA dataset is a massively multilingual instruction fine-tuning dataset. AYA could provide further means of realising the demonstrated benefits of multilingual training \cite{nieMultilingualLargeLanguage2024}.

\section{Conclusion}
Increased multilingual capabilities do not guarantee improved cultural alignment in Large Language Models. Through systematic comparison of three model families---Gemma, OLMo, and OpenAI's GPTs---we find that the relationship between improvements in multilingual capability and cultural alignment is complex. While some languages show clear improvements in alignment with increased model capabilities (e.g., Danish), others exhibit inconsistent patterns, suggesting that cultural alignment does not automatically follow gains in multilingual capabilities. Our distribution-matching methodology using World Values Survey data enabled the detection of these nuanced patterns across languages and cultural contexts.

We also find that, contrary to popular discourse, LLMs do not exhibit US-centric bias across all languages; in Danish and Dutch, they align more closely with the values of Denmark and the Netherlands, respectively, than with the US. This fits with the hypothesis that more culturally uniform data leads to less US-centric bias. Both English and Portuguese are spoken in multiple countries, whereas Dutch and Danish are predominantly spoken in one. To further validate this claim, future work could include other multi-cultural languages (like Spanish or Swahili) and monocultural languages (like Japanese)---especially with a wider geographical reach to preclude European bias.

Our findings highlight that improving cultural alignment requires dedicated effort beyond general capability scaling. Future work should focus on developing techniques that can better handle alignment with distributions of cultural values rather than single points, while ensuring meaningful participation from diverse communities in LLM development. As these models continue to reach wider audiences spanning many geographic and cultural regions, achieving robust cultural alignment becomes increasingly crucial for equitable deployment.

\section*{Acknowledgements}
We are thankful for the helpful feedback from the anonymous reviewers. We also thank Shiri Dori-Hacohen, Daniel Hershcovich, and others for helpful discussions throughout the project. For compute support, the project used the Microsoft Azure Accelerating Foundation Model Research Grant. 
This work was supported in part by the Engineering and Physical Sciences Research Council [grant number EP/X028909/1].


\bibliographystyle{acl_natbib}
\bibliography{references}
\end{document}